\documentclass[conference]{IEEEtran}

\usepackage{color}
\usepackage{amsmath,amssymb,graphicx,hyperref}
\usepackage{algorithm}
\usepackage{bm}
\usepackage{verbatim}
\usepackage{algpseudocode}

\newcommand{\bw}{\mathbf{w}}

\title{Conflict-Aware Client Selection for Multi-Server Federated Learning}

\author{
    \IEEEauthorblockN{
        Mingwei Hong\IEEEauthorrefmark{1},
        Zheng Lin\IEEEauthorrefmark{2},
        Zehang Lin\IEEEauthorrefmark{1},
        Lin Li\IEEEauthorrefmark{1},
        Miao Yang\IEEEauthorrefmark{1},
        Xia Du\IEEEauthorrefmark{1}, \\
        Zihan Fang\IEEEauthorrefmark{3},
        Zhaolu Kang\IEEEauthorrefmark{4},
        Dianxin Luan\IEEEauthorrefmark{5} and
        Shunzhi Zhu\IEEEauthorrefmark{1}
    }
    \IEEEauthorblockA{\IEEEauthorrefmark{1}School of Computer and Information Engineering, Xiamen University of Technology, Xiamen, China}
    \IEEEauthorblockA{\IEEEauthorrefmark{2}Department of Electrical and Electronic Engineering, The University of Hong Kong, Hong Kong, China}
    \IEEEauthorblockA{\IEEEauthorrefmark{3}Department of Computer Science, City University of Hong Kong, Hong Kong, China}
    \IEEEauthorblockA{\IEEEauthorrefmark{4}School of Software \& Microelectronics, Peking University, Beijing, China}
    \IEEEauthorblockA{\IEEEauthorrefmark{5}Institute for Imaging, Data and Communications, University of Edinburgh, UK}
}

\begin{document}

\maketitle

\begin{abstract}
Federated learning (FL) has emerged as a promising distributed machine learning (ML) that enables collaborative model training across clients without exposing raw data, thereby preserving user privacy and reducing communication costs. Despite these benefits, traditional single-server FL suffers from high communication latency due to the aggregation of models from a large number of clients. While multi-server FL distributes workloads across edge servers, overlapping client coverage and uncoordinated selection often lead to resource contention, causing bandwidth conflicts and training failures. To address these limitations, we propose a decentralized reinforcement learning with conflict risk prediction, named RL-CRP, to optimize client selection in multi-server FL systems. Specifically, each server estimates the likelihood of client selection conflicts using a categorical hidden Markov model based on its sparse historical client selection sequence. Then, a fairness-aware reward mechanism is incorporated to promote long-term client participation for minimizing training latency and resource contention. Extensive experiments demonstrate that the proposed RL-CRP framework effectively reduces inter-server conflicts and significantly improves training efficiency in terms of convergence speed and communication cost.
\end{abstract}
\begin{IEEEkeywords}
Multi-server federated learning, conflict selection, reinforcement learning, fairness consideration.
\end{IEEEkeywords}

\maketitle

\section{Introduction}
\label{sec:intro}
Federated learning (FL)~\cite{hu2024accelerating,zhang2025lcfed} is a distributed machine learning paradigm that facilitates collaborative model training across distributed clients without exposing raw data  \cite{NetoHDMF23,lin2024adaptsfl,KonecnyMRR16,lin2025hierarchical,fang2026hfedmoe}. By uploading model updates rather than private data to a central server for aggregation, FL significantly enhances privacy preservation and mitigates the risks associated with centralized data collection,  making it particularly advantageous in regulation-intensive domains such as healthcare and finance \cite{LongT0Z20,lin2023pushing,ZhangXBYLG21}. Additionally, FL improves model generalizability by leveraging heterogeneous data sources while reducing storage and communication costs. Conventional FL frameworks predominantly adopt a single-server architecture~\cite{lin2025sl}, wherein a central server orchestrates client selection and model aggregation. However, this centralized design often underutilizes the distributed computing and communication resources of edge networks \cite{LiSTS20}. To address this issue, multi-server FL architectures have been proposed, wherein multiple edge servers collaboratively manage subsets of clients to enhance scalability and reduce communication latency \cite{AbadOGE20}. However, clients often reside within overlapping regions of multiple servers, and uncoordinated client selection can lead to resource contention, resulting in training timeouts due to limited bandwidth and computing resources.

To address the communication bottleneck in multi-server FL, several approaches have been proposed. Clustered FL groups clients into disjoint groups, each training an independent model to reduce inter-group communication, but repeated clustering across rounds introduces substantial computational overhead and coordination complexity~\cite{SattlerMS21}. Hierarchical FL employs edge servers for intermediate aggregation before global updates, alleviating cloud burden but suffering from excessive latency overhead due to edge-cloud synchronization delays~\cite{Liu0SL20}. Over-the-air computation exploits the superposition property of wireless channels to enable simultaneous uplink transmission, offering improved spectrum efficiency. However, analog implementations are highly susceptible to noise, while digital variants impose significant encoding and decoding complexity~\cite{Han2025TMC}. From a theoretical perspective, MS-FedAvg enables indirect model sharing through overlapping client regions and provides convergence guarantees under non-convex objectives, but its practicality is hindered by reliance on client availability in overlapping regions and idealized bandwidth conditions~\cite{Qu2023TMC}.
Despite these advances, existing multi-server FL methods primarily focus on communication efficiency, while largely overlooking contention resolution and fairness in client selection. As a result, they exhibit high sensitivity to dynamic and heterogeneous network environments, limiting their robustness and scalability in practical deployments.

To address these issues, we propose a decentralized reinforcement learning with conflict risk prediction (RL-CRP) for client selection in multi-server FL. 
Each server employs conflict risk prediction (CRP), utilizing a categorical hidden Markov model (HMM) to complete outdated historical sequences for clients, and estimate the individual conflict risk based on the selection sequence of clients.
Then, each server independently selects clients using only local observations based on deep reinforcement learning (RL).
We design a novel fairness-aware reward to encourage sustained participation, thereby improving the training performance.
The main contributions of this work are summarized as follows:
\begin{itemize}
\item We propose a CRP mechanism based on categorical based on HMM, which accurately estimates the probability of client selection conflicts from sparse and outdated client historical data.
\item We develop a decentralized RL framework that enables servers to perform conflict-aware and fairness-driven client selection, significantly reducing inter-server contention and improving training efficiency.
\item We conduct extensive experiments to validate the effectiveness of the proposed RL-CRP, demonstrating its superiority in reducing conflicts and improving convergence.
\end{itemize}

The rest of the paper is organized as follows. Section~\ref{sec:system model} elaborates on the system model of multi-server FL. 
Section~\ref{sec:algorithm} presents the proposed RL-CPR algorithm. 
Section~\ref{sec:experiment} introduces the performance evaluation. Finally, conclusions are presented in Section~\ref{sec:conclusion}.

\section{SYSTEM MODEL}
\label{sec:system model}

\begin{figure}[t]
  \centering
  \centerline{\includegraphics[width=7.5cm]{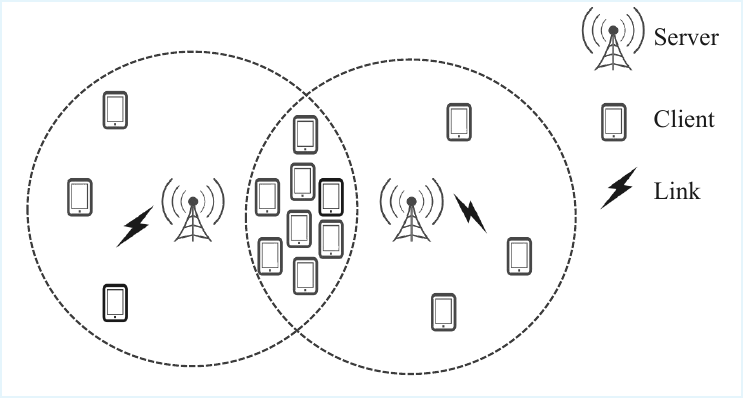}}
  \vspace{-4mm}
  \caption{Illustration of multi-server FL.}
  \label{fig:system_model}
  \vspace{-2mm}
\end{figure}

We consider a multi-server FL system consisting of $M$ servers and $N$ clients. 
Let $\mathcal{M} = \{1, \dots, M\}$ denote the set of servers and $\mathcal{N} = \{1, \dots, N\}$ the set of clients. 
As shown in Fig.~\ref{fig:system_model}, each server $m \in \mathcal{M}$ is associated with a disjoint subset of clients $\mathcal{N}_m \subseteq \mathcal{N}$ and maintains its own local model $\mathbf{w}_m$. The local objective function for server $m$ is defined as:
\begin{align}
    F_m(\mathbf{w}_m) = \sum_{k \in \mathcal{N}_m} \frac{n_k}{N_m} f_k(\mathbf{w}_m),
\end{align}
where $n_k$ denotes the number of data samples at client $k$, and $N_m = \sum_{k \in \mathcal{N}_m} n_k$ is the total number of samples managed by server $m$.

Therefore, the global objective of multi-server FL could be formulated as a multi-task problem across all servers:
\begin{equation}
\min_{\{\bw_m\}_{m=1}^M}\; F(\{\bw_m\}) \;=\; \sum_{m=1}^M \frac{N_m}{N}\, F_m(\bw_m),
\label{eq:global-obj}
\end{equation}

In practical deployments, due to constraints on communication bandwidth, each server can only select a fixed number of $S$ clients per communication round. However, as some clients are located within the overlapping coverage areas of multiple servers, they may be simultaneously selected by more than one server. This contention for shared clients often leads to resource conflicts, causing delayed or failed model updates that degrade communication efficiency and prolong the overall training time.
To mitigate such conflicts, we propose a decentralized RL algorithm that predicts client conflict probabilities from historical information,and then incorporates a fairness-aware reward to ensure balanced participation.

\section{Conflict-Aware Client Selection in Multi-Server Federated Learning}
\label{sec:algorithm}

We propose a multi-server FL system that features decentralized client selection, effectively avoiding inter-server conflicts while ensuring long-term fairness in client participation. The framework is designed by two parts: conflict risk prediction and decentralized RL with fairness guarantee.

\subsection{Conflict Risk Prediction with Sparse Historical Sequence}
\label{sec:conflict_prediction}
To accurately identify suitable client selection subsets for each server, it is essential to predict the conflict risk of individual clients. In this work, we leverage the historical selection sequence of each client to estimate its probability of being selected. However, in practical FL scenarios, it is challenge to obtain the complete selection history of all clients. Without loss of generality, we assume that if a client was last selected at time $t - d$ (where $d > 1$), the server only has access to its historical record up to time $t - d$. Consequently, accurately predicting conflict risks at the current time step $t$ requires addressing the inherent sparsity and incompleteness of historical client selection sequences.

We employ categorical HMM to derive the conflict risk of each client \cite{maruotti2012mixed}.
The categorical HMM is parameterized with $K$ hidden states $h_t \in \{1, 2, \ldots, K\}$, and let $V$ denote the number of possible observation categories (e.g., $V=2$ for binary conflict indicators).
The model is parameterized by: a state transition matrix $\mathbf{A}$ where $A_{ij} = P(h_{t+1} = j \mid h_t = i)$, an emission probability matrix $\mathbf{B} \in \mathbb{R}^{K \times V}$ where $B_j(v) = P(o_t = v \mid h_t = j)$ for a specific observation value $v$, an initial state distribution $\bm{\pi}$, and the complete parameter set $\lambda = (\mathbf{A}, \mathbf{B}, \bm{\pi})$.
Here, the observation $o_t$ is categorical as $o_t = 1$ indicates a conflict event, and $o_t = 0$ denotes normal operation.

For a client with an observed sequence $O_i{[0:t-d]}$ where $d > 1$, we aim to predict the conflict probability at time $t$. We first compute the forward probability $\alpha_t(i)$ via the forward algorithm: 
\begin{align}
    \alpha_t(i) = P(o_1, \ldots, o_t, h_t = i \mid \lambda).
    \label{equ:alpha}
\end{align}
The posterior distribution over hidden states at time $t-d$ is then given by:
\begin{align}
\gamma_{t-d}(i)\!=\!P(h_{t-d}\!=\!i \mid O_i{[0:t-d]})\!=\!\frac{\alpha_{t-d}(i)}{\sum_{j=1}^K \alpha_{t-d}(j)}.
\end{align}
Then, we propagate this distribution to time $t - 1$ via the transition matrix:
\begin{align}
\bm{\gamma}_{t - 1} = \bm{\gamma}_{t - d} \cdot \mathbf{A}^{d - 1}.
\end{align}
Finally, the conflict probability at time $t$ is computed as:
\begin{align}
p_i^{(t)} = P(o_t = 1 \mid O_i{[0:t-d]}) = \bm{\gamma}_{t-1} \cdot \mathbf{A} \cdot \mathbf{B}{[:, 1]},
\label{equ:p}
\end{align}
where $\mathbf{B}{[:, 1]}$ denotes the emission probability vector for conflict events ($o_t = 1$).

To update the model parameters as new observations become available, we use an incremental version of the Baum–Welch algorithm~\cite{baum1970maximization}. 
Given a new observation sequence $O^{(n+1)}$, and current parameters $\lambda^{(n)}$, we first compute the backward probability $\beta_t^{(n)}(i)$ as:
\begin{align}
    \beta_t(i) = P(o_{t+1}^{(n)}, \ldots, o_T^{(n)} \mid h_t = i, \lambda^{(n)}).
    \label{equ:beta}
\end{align}

The probability of a transition from state $i$ to $j$ at time $t$ is given by:
\begin{align}
\xi_t^{(n+1)}(i,j)
&= P(h_t=i, h_{t+1}=j \mid O^{(n+1)}, \lambda^{(n)}) \nonumber\\
&= \frac{\alpha_t^{(n)}(i)\, A_{ij}^{(n)}\, B_j^{(n)}(o_{t+1})\, \beta_{t+1}^{(n)}(j)}
{P(O^{(n+1)}\mid\lambda^{(n)})},
\end{align}
and the posterior probability of state $i$ at time $t$ is
\begin{align}
\gamma_t^{(n+1)}(i)
&= P(h_t = i \mid O^{(n+1)}, \lambda^{(n)}) \nonumber\\
&= \frac{\alpha_t^{(n)}(i) \beta_t^{(n)}(i)}
{P(O^{(n+1)}\mid\lambda^{(n)})},
\end{align}
where $P(O^{(n+1)}\mid\lambda^{(n)})=\sum_{k=1}^{K}\alpha_T^{(n)}(k)$ is the probability of the entire sequence $O^{n+1}$ under model $\lambda^{(n)}$.

Then, we update parameters via exponential moving average:
\begin{align}
A_{ij}^{(n+1)} &= \frac{\sum_{t=1}^{T-1} \xi_t^{(n+1)}(i,j)}{\sum_{t=1}^{T-1} \gamma_t^{(n+1)}(i)}, \\
B_{j}^{(n+1)}(v)&=\frac{\sum_{t=1}^{T}\gamma_t^{(n+1)}(j)\mathbb{I}(o_t^{(n+1)}=v)}{\sum_{t=1}^T\gamma_t^{(n+1)}(j)}, \\
\pi^{(n+1)}_i &= \gamma_1^{(n+1)}(i),\label{equ:update}
\end{align}
where $\mathbb{I}(\cdot)$ is the indicator function that equals $1$ when the condition is true and $0$ otherwise.

\subsection{Fairness-aware Client Selection based on RL}
Within the decentralized RL framework, each agent must independently make client selection decisions based on its local observations and predictions. To formalize this process, we define the following Markov Decision Process (MDP) components for each agent, which will be solved using the Soft Actor-Critic (SAC) algorithm \cite{HaarnojaZHTHKZGAL18}.

\textbf{State Space.}
The state of server $m$ is defined as:
\begin{equation}
s_m = \left( \mathcal{L}_m, \mathcal{P}_m \right),
\end{equation}
where $\mathcal{L}_m = \{L_m^k\}_{k\in\mathcal{N}_m}$ represents the observed upload latency vector of all clients associated with server $m$ under the current communication environment, and $\mathcal{P}_m = \{p_m^k\}_{k\in\mathcal{N}_m}$ denotes the conflict probability vector of these clients, as estimated in Section~\ref{sec:conflict_prediction}.

\textbf{Action Space.}
Each server selects a subset of clients to participate in the current FL round. The action of server $m$ is defined as:
\begin{equation}
a_m = \left( c_1, c_2, \dots, c_S \right).
\end{equation}

\textbf{Reward Function.}
To jointly optimize communication efficiency, conflict avoidance, and client fairness, we design the following reward function for server $m$:
\begin{align}
    r_m = -L_m - C_m + \alpha \cdot f,
    \label{equ:reward}
\end{align}
where \( L_i \) is the total communication latency incurred by the overall training latency of server \( m \), which is determined by the client with worst communication and computation,
\( C_i \) denotes a penalty term quantifying the cost of conflicts or timeouts caused by the current selection, \( f \) represents a fairness metric defined as:
\begin{align}
    f = \tanh\left( \frac{\mu}{\delta + \epsilon} \right),
\end{align}
where $\mu$ is the average number of training rounds per client, \( \delta \) is the standard deviation, and \( \epsilon \) is a small constant to avoid division by zero. This term approximates the inverse of the coefficient of variation, promoting equitable client participation, \( \alpha \) is a weighting coefficient that balances the fairness objective against latency and conflict costs.

We employ SAC algorithm to learn a policy that maximizes the cumulative discounted reward \cite{HaarnojaZAL18}. 
Each server maintains a policy network \( \pi_{\phi}(a_m \vert s_m) \) that outputs a distribution over actions given the state, two Q-networks \( Q_{\theta_1}(s_m, a_m) \), \( Q_{\theta_2}(s_m, a_m) \) to reduce overestimation bias, a target Q-network for each Q-network to stabilize training.
The Q-functions are updated by minimizing the Bellman error:
\begin{align}
\hspace{-2mm}&\mathcal{L}(\theta_i) = \mathbb{E}_{(s_m, a_m, r_m, s'_m) \sim \mathcal{D}} \Big[  Q_{\theta_i}(s_m, a_m)-\Big( r_m\nonumber\\
& + \gamma \Big( \min_{j=1,2} Q_{\bar{\theta}j}(s'_m, a'_m) - \eta \log \pi\phi(a'_m | s'_m)  \Big) \Big)^2 \Big],
\end{align}
where $\eta$ is the temperature parameter and $\mathcal{D}$ is the replay buffer.
The policy parameters $\phi$ are updated to maximize the expected return and entropy:
\begin{align}
\!\!\mathcal{L}(\phi)\!=\!\mathbb{E}_{s_m \sim \mathcal{D}}\!\left[\!\eta \log \pi\phi(a_m | s_m)\!-\!\min_{j=1,2} Q_{\theta_j}(s_m, a_m)\!\right],
\end{align}
where the temperature parameter $\eta$ is automatically adjusted to maintain a target entropy level:
\begin{align}
\mathcal{L}(\eta) = \mathbb{E}_{s \sim \mathcal{D}} \left[ -\eta \left( \log \pi\phi(a_m | s_m) + \mathcal{H} \right) \right],
\end{align}
where $\mathcal{H}$ is the target entropy\cite{HaarnojaZAL18}.
During execution, each server uses its policy network to select actions based on local states, enabling scalable and independent decision-making without inter-server communication.

\begin{algorithm}[t]
\caption{Decentralized RL with Conflict Risk Prediction}
\label{alg}
\begin{algorithmic}[1]
\State \textbf{Initialize}: SAC policy $\pi_i$, value networks $Q_{i1},Q_{i2}$, 
targets $Q'_{i1},Q'_{i2}$, conflict net $W$, HMM parameters $\lambda$, total bandwidth $BW$.
\For{$t=1,\dots, T$}
  \For{each server $m=1, \dots, M$}
    \State Observe latency $L_i$.
    \State Gets conflict probability $\mathcal{P}_m$ by equation (\ref{equ:alpha})-(\ref{equ:p}).
    \State Obtain selection subset via $a_m \sim \pi_{\phi}(s_m)$, allocate bandwidth to the selected clients, and transmit global model $\mathbf{w}_m$.
  \EndFor
  \State Each client trains and upload model to the corresponding server for model aggregation.
  \State Compute reward with equation (\ref{equ:reward}).
  \State Update $Q_{\theta_1}(s_m, a_m), Q_{\theta_2}(s_m, a_m), \pi_{\phi)}$.
  \State Update conflict prediction parameter $\lambda$ via (\ref{equ:beta})-(\ref{equ:update}).
\EndFor
\end{algorithmic}
\end{algorithm}

After selecting the devices, we allocate channel bandwidth using a water-filling algorithm \cite{NishioY19}. Clients are sorted in descending order of channel quality. Bandwidth is assigned sequentially from the best channel condition until the total available bandwidth $M$ is exhausted. The process terminates when the remaining bandwidth cannot satisfy the next client. This approach prioritizes clients with the strongest channels under the bandwidth constraint. The overall RL-CRP procedure is summarized in Algorithm~\ref{alg}.

\section{Numerical Results}
\label{sec:experiment}
\subsection {Experimental Setup}
\textbf {Dataset and Model.} We evaluate the proposed RL-CRP algorithm on the CIFAR-10 dataset \cite{krizhevsky2009learning}. 
Experiments are conducted under both IID and non-IID data distributions. For the non-IID setting, client data is partitioned using a Dirichlet distribution with parameter $\eta = 0.1$. 
The global model of FL is implemented as convolutional neural networks with three convolutional layers. Each client trains its local model using mini-batch stochastic gradient descent (SGD) with a batch size of 16, for 5 local epochs, and a learning rate of 0.005.
Note that we consider a sustained training scenario in which the RL-CRP algorithm is designed to train multiple FL models sequentially.

\textbf{Communication Environment.}
The communication environment comprises two servers deployed at different locations, each with a 1 km coverage radius. Among a total of $N = 50$ clients, 40 are located within the servers’ coverage. The bandwidth is set to 100 MHz, the maximum latency $L_{\text{max}}$ to 40 seconds, and the fairness weight $\alpha$ to 100.

\textbf{Benchmarks.}
We compare RL-CRP against three baseline methods: (i) Fedavg\cite{McMahanMRHA17}, which randomly selects clients; (ii) RL-CRP without the fairness component (abbreviated as RL-CRP w/o fairness); and (iii) ENSAC \cite{DengWLCGHTHLL24}: an entropy-normalized soft actor-critic algorithm for computation offloading in FL.

\subsection {Superb performance of RL-CRP}


\begin{figure}
\begin{minipage}[b]{0.48\linewidth}
  \centering
  \centerline{\includegraphics[width=4.5cm]{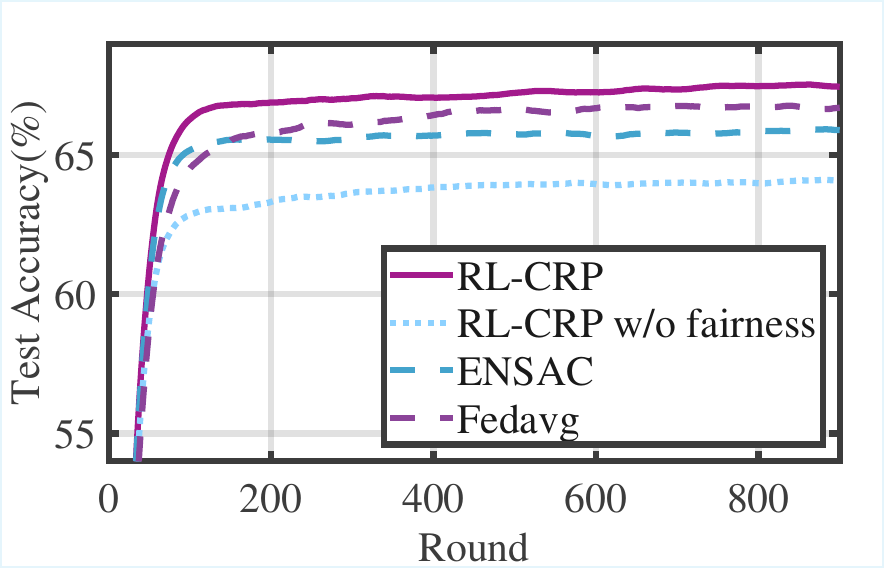}}

      \centerline{(a) IID setting }\medskip  
  
\end{minipage}
\hfill
\begin{minipage}[b]{0.48\linewidth}
  \centering
  \centerline{\includegraphics[width=4.5cm]{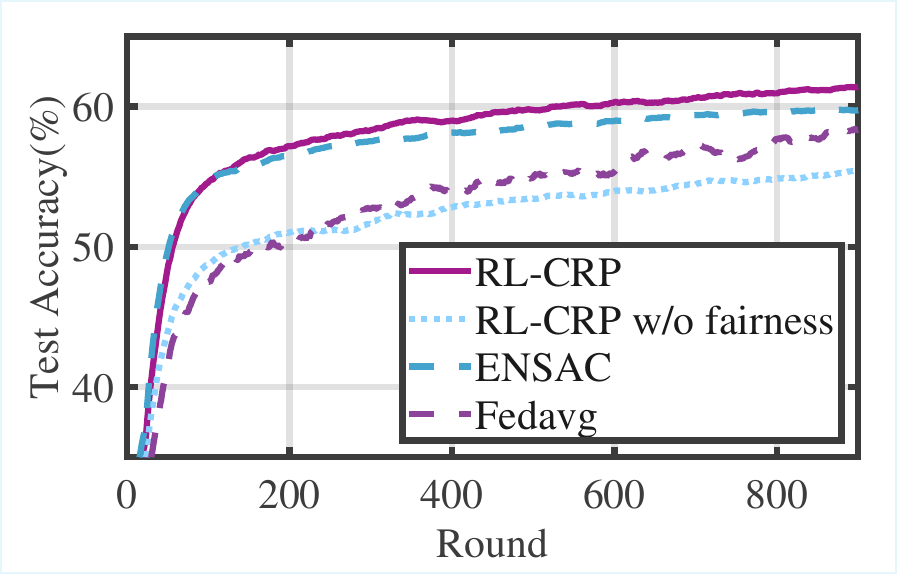}}

  \centerline{(b) non-IID setting }\medskip  
  
\end{minipage}
\vspace{-4mm}
\caption{Test accuracy of baseline algorithms.}

\label{fig:iid}
\end{figure}

\begin{figure}
\begin{minipage}[b]{0.48\linewidth}
  \centering
\centerline{\includegraphics[width=4.2cm]{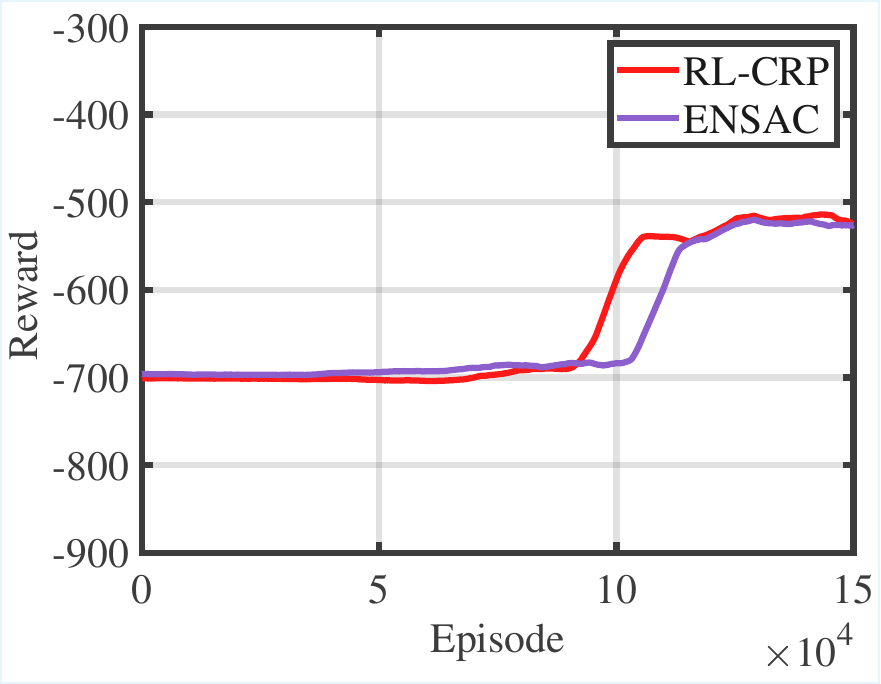}}
      \centerline{(a) Reward of 2 servers }\medskip  
      
\end{minipage}
\hfill
\begin{minipage}[b]{0.48\linewidth}
  \centering
 \centerline{\includegraphics[width=4.4cm]{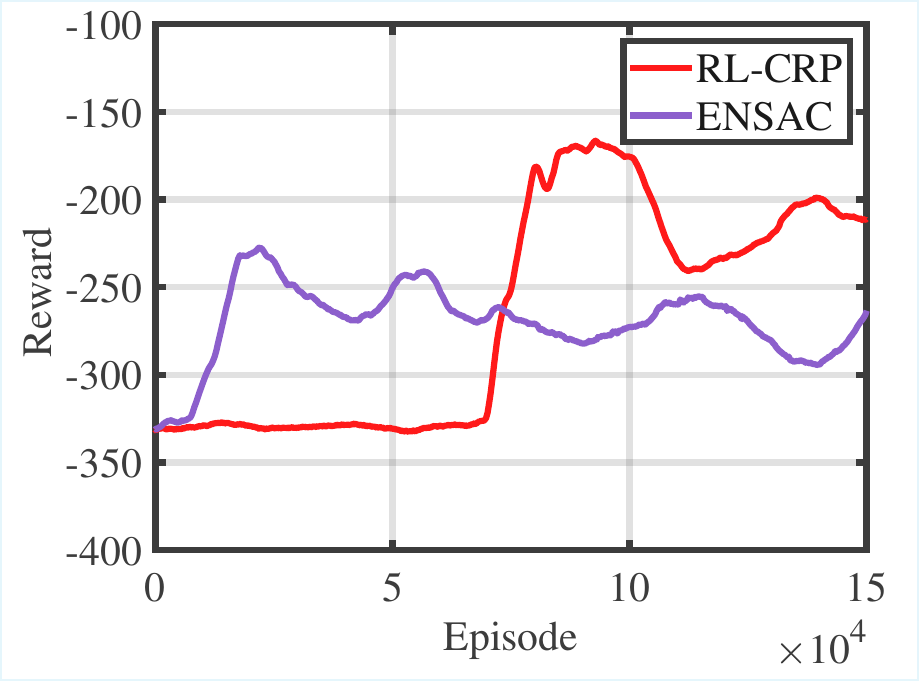}}
  \centerline{(b) Reward of 4 servers }\medskip  
\end{minipage}
\vspace{-4mm}
\caption{Reward performance of baseline algorithms with different number of servers.}
\label{fig:iidacc}
\vspace{-4mm}
\end{figure}

Fig.~\ref{fig:iid} presents the test accuracy  performance of the compared algorithms.
As shown in Fig.~\ref{fig:iid} (a) and (b), RL-CRP achieves the highest accuracy among all the benchmarks, reaching convergence values of 67.68\% under the IID setting, and 61.32\% under the non-IID setting.
The incorporation of fairness into the reward function encourages the inclusion of previously underrepresented clients, which ultimately enhances model accuracy.
However, we observe that the variant of RL-CRP without fairness achieves considerably lower performance.
This is because the absence of fairness results in reduced participation from clients with poor communication or computational capacity, leading to noticeably inferior training accuracy.
Besides, in Fig.~\ref{fig:iid} (b), we can observe that all algorithms exhibit a slight performance degradation compared to the IID setting. This decline can be attributed to the statistical heterogeneity inherent in non-IID data distributions, which increases the difficulty of model convergence in multi-server federated learning. Nevertheless, RL-CRP mitigates this issue more effectively through its conflict reduction and fairness-aware client selection mechanisms.


Fig.~\ref{fig:iidacc} illustrates the reward convergence performance of the proposed RL-CRP and the baseline ENSAC algorithm. As depicted in Fig.~\ref{fig:iidacc} (a), RL-CRP achieves significantly faster reward convergence than ENSAC, which demonstrates its effectiveness in reducing client-server conflicts through historical sequence modeling. Fig.~\ref{fig:iidacc} (b) further examines the impact of system scale by increasing the number of servers. With more servers, conflict probability increases, leading to longer convergence time for all algorithms. Nevertheless, RL-CRP achieve the better reward performance, while ENSAC converges to a significantly lower final reward. These results highlight the robustness and scalability of RL-CRP in multi-server FL systems, underscoring the importance of conflict prediction for sustaining training efficiency under expanding system scales.

\begin{figure}[]
\begin{minipage}[b]{0.45\linewidth}
  \centering
  \centerline{\includegraphics[width=4.4cm]{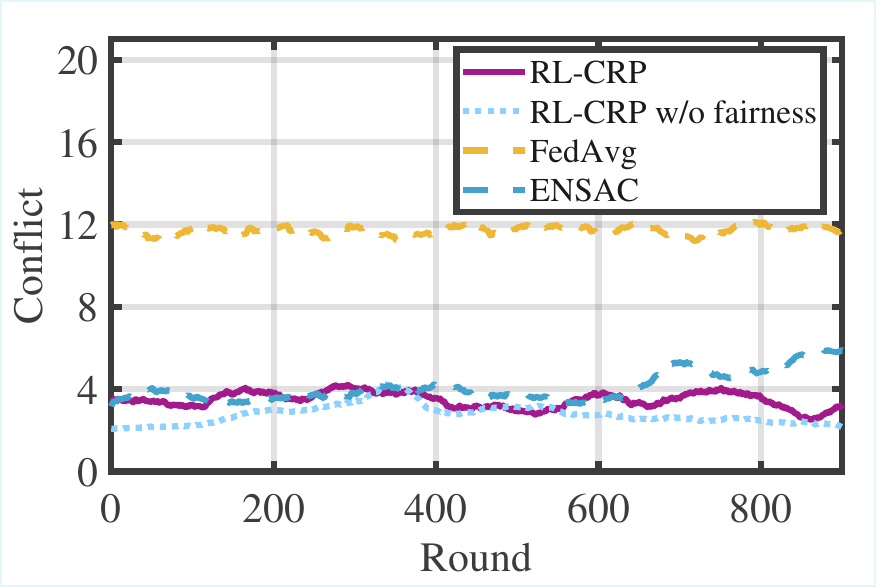}}
  \centerline{(a) Conflict of 2 servers    }\medskip  
\end{minipage}
\hfill
\begin{minipage}[b]{0.50\linewidth}
  \centering
  \centerline{\includegraphics[width=4.5cm]{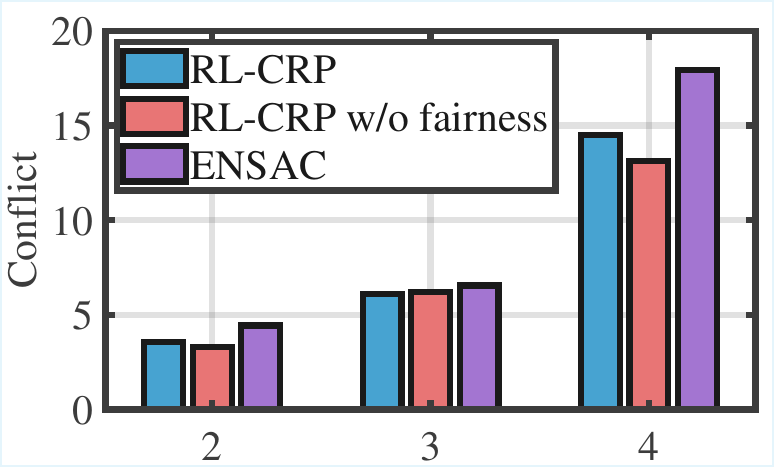}}
  \vspace{0.3cm}
  \centerline{(b) Conflict with increased servers  }\medskip  
\end{minipage}
\vspace{-4mm}
\caption{Conflict performance versus different number of servers.}
\label{fig:multi}
\vspace{-4mm}
\end{figure}

Fig.~\ref{fig:multi} (a) illustrates the number of conflicts occurring during FL training for 2 servers when the baseline algorithms are stable.
It can be observed that RL-CRP achieves fewer conflicts compared to both ENSAC and FedAvg during training, demonstrating the effectiveness of the CRP module.
Furthermore, RL-CRP without fairness is observed to incur the fewest conflicts.
This suggests that incorporating fairness tends to increase client selection conflicts in multi-server FL.
The results reveal a trade-off between conflict minimization and fairness, suggesting that while fairness improves global accuracy, it may come at the cost of increased conflicts.



Fig.~\ref{fig:multi} (b) depicts the conflict performance of the baseline algorithms as the number of servers varies.
As shown in Fig.~\ref{fig:multi} (b), the conflict count shows a consistent upward trend across all considered algorithms as the number of servers increases, mainly because the expansion of server deployment leads to larger overlapping coverage areas and consequently intensifies the competition for limited client resources.
Despite this trend, both RL-CRP and its fairness-ablated variant consistently achieve fewer conflicts than ENSAC, highlighting the robustness and scalability of the conflict prediction mechanism embedded in RL-CRP.
These results suggest that conflict avoidance remains effective even as system scale expands, indicating the utility of historical sequence modeling for coordination in multi-server federated learning.



\section{CONCLUSION}

\label{sec:conclusion}
In this paper, we have proposed a RL-based client selection framework with conflict risk prediction, termed RL-CRP, to mitigate inter-server contention and enhance training efficiency in multi-server RL. RL-CRP first employs categorical HMM to predict client selection conflicts based on sparse historical sequences.
Then, RL-CRP incorporates a fairness-aware reward mechanism that promotes balanced client participation while minimizing latency and resource competition. 
Extensive experimental results demonstrate that RL-CRP significantly reduces conflict occurrences and improves convergence speed compared to existing benchmarks. As a potential future direction, we are looking forward to extending our method to improve the performance of various applications such as satellite netwotks~\cite{zhao2024leo,yuan2023graph,lin2025esl}, distributed learning system~\cite{zhang2024fedac,zhan2025prism}, and LLM system~\cite{tang2024merit,duan2025leed,lin2025hsplitlora}.

\bibliographystyle{IEEEbib}
\bibliography{strings,refs}

\end{document}